\documentclass[10pt,twocolumn,letterpaper]{article}

\usepackage[pagenumbers]{cvpr} 

\usepackage{graphicx}
\usepackage{amsmath}
\usepackage{amssymb}
\usepackage{algorithm}
\usepackage{algpseudocode}
\usepackage{booktabs}
\usepackage{blindtext}
\usepackage{overpic}
\usepackage{xspace}
\usepackage{colortbl}
\usepackage{xcolor}
\usepackage{pifont}
\usepackage{lipsum}
\usepackage{tikz}
\usepackage{microtype}
\usetikzlibrary{shadows.blur}

\definecolor{turquoise}{cmyk}{0.65,0,0.1,0.3}
\definecolor{purple}{rgb}{0.65,0,0.65}
\definecolor{dark_green}{rgb}{0, 0.5, 0}
\definecolor{orange}{rgb}{0.8, 0.6, 0.2}
\definecolor{red}{rgb}{0.8, 0.2, 0.2}
\definecolor{darkred}{rgb}{0.6, 0.1, 0.05}
\definecolor{blueish}{rgb}{0.0, 0.3, .6}
\definecolor{light_gray}{rgb}{0.7, 0.7, .7}
\definecolor{pink}{rgb}{0.9, 0, 0.6}
\definecolor{greyblue}{rgb}{0.25, 0.25, 1}
\definecolor{teal}{rgb}{0.0, 0.4, 0.4}
\definecolor{chocolate}{rgb}{1.0, 0.4, 0.0}

\definecolor{figred}{rgb}{0.8352941176470589, 0.24313725490196078, 0.30980392156862746}
\definecolor{figgreen}{rgb}{0.3070588235294118, 0.5498039215686275, 0.2749019607843137}
\definecolor{figblue}{rgb}{0.19607843137254902, 0.5333333333333333, 0.7411764705882353}

\newcommand{\TODO}[1]{\textbf{\color{red}[TODO: #1]}}
\renewcommand{\TODO}[1]{}

\usepackage{soul}
\setuldepth{foobar}

\newcommand{\dropshadow}[1]{%
    \begin{tikzpicture}
        \node[draw=none, fill=none, blur shadow={shadow blur steps=50, shadow xshift=0pt, shadow yshift=0pt}, inner sep=0pt] 
        {#1};
    \end{tikzpicture}%
}

\renewcommand{\paragraph}[1]{\vspace{.5em}\noindent\textbf{#1}.}

\DeclareMathOperator*{\argmin}{arg\,min}

\newcommand{\expect}{\mathbb{E}}
\newcommand{\losst}[1]{\mathcal{L}_\text{#1}}

\newcommand{\calU}{\mathcal{U}}
\newcommand{\real}{\mathbb{R}}
\newcommand{\ray}{\mathbf{r}}
\newcommand{\radiance}{\mathbf{c}}

\newcommand{\density}{\sigma}
\newcommand{\transmittance}{T}

\newcommand{\primal}{\mathbf{p}}
\newcommand{\verts}{\mathcal{V}}
\newcommand{\cell}{\mathbf{c}}
\newcommand{\cells}{\mathcal{C}}

\newcommand{\mipnerf}{MipNeRF~360~\cite{mipnerf360}\xspace}

\newcommand{\deepblend}{Deep~Blending~\cite{deepblending}\xspace}

\definecolor{color1}{rgb}{0.9, 0.65, 0.65}
\definecolor{color2}{rgb}{0.95, 0.8, 0.8}
\definecolor{color3}{rgb}{1.0, 0.9, 0.9}

\newcommand{\cmark}{\textcolor{green}{\ding{51}}} 
\newcommand{\xmark}{\textcolor{red}{\ding{55}}}   

\def\paperID{XXXX} 
\def\confName{CVPR}
\def\confYear{2025}

\title{Radiant Foam: Real-Time Differentiable Ray Tracing}

\author{
Shrisudhan Govindarajan*$^{1}$, 
Daniel Rebain*$^{2}$, 
Kwang Moo Yi$^{2}$, 
Andrea Tagliasacchi$^{1,3,4}$
\\[.5em]
$^{1}$Simon Fraser University,
$^{2}$University of British Columbia,
\\
$^{3}$University of Toronto,
$^{4}$Google DeepMind, 
$^{*}$equal contributions
\\[.5em]
\url{radfoam.github.io}
\vspace{-1em}}

\begin{document}
\twocolumn[{%
\renewcommand\twocolumn[1][]{#1}%
\maketitle
\dropshadow{\includegraphics[height=.28\linewidth]{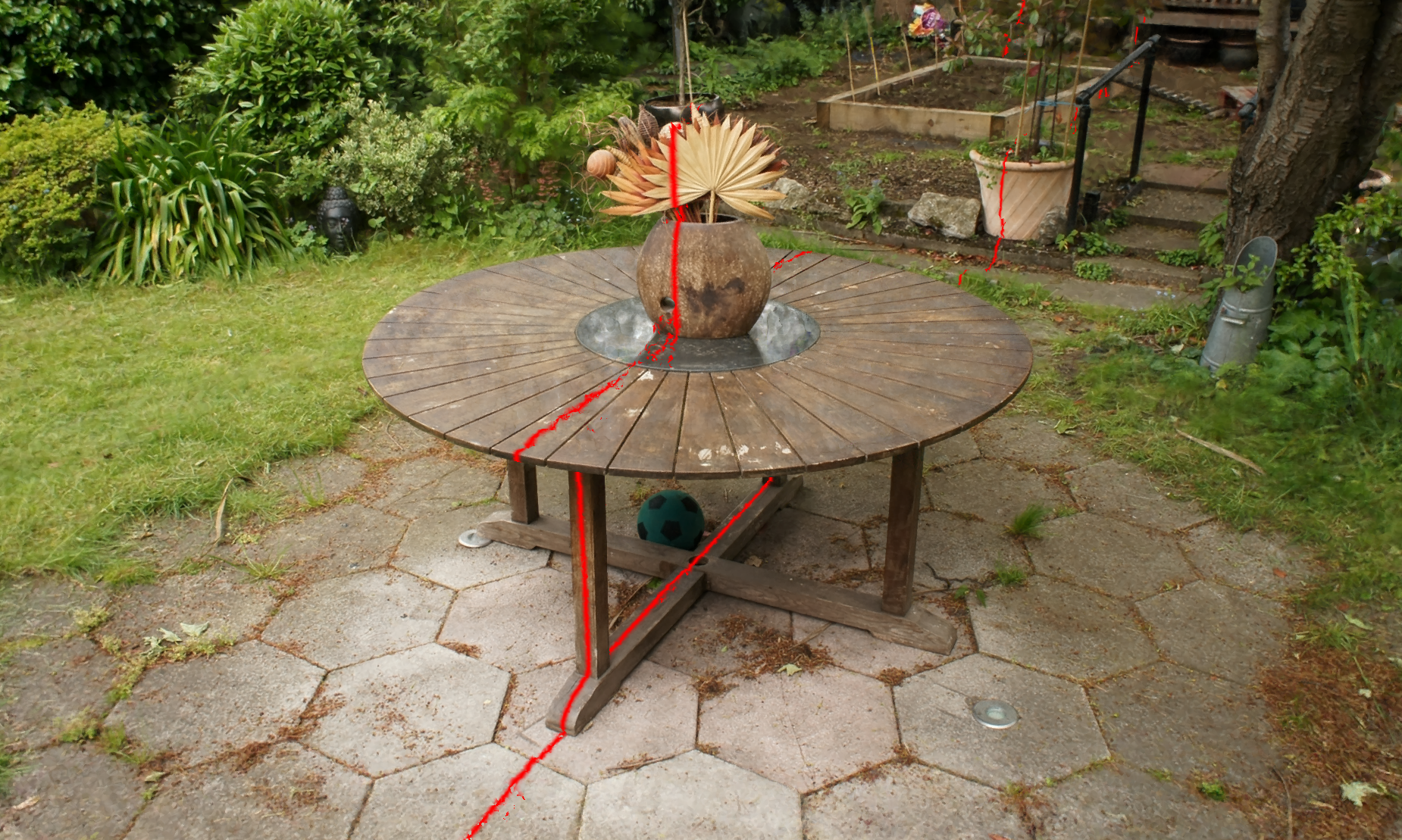}}
\hfill
\dropshadow{\includegraphics[height=.28\linewidth]{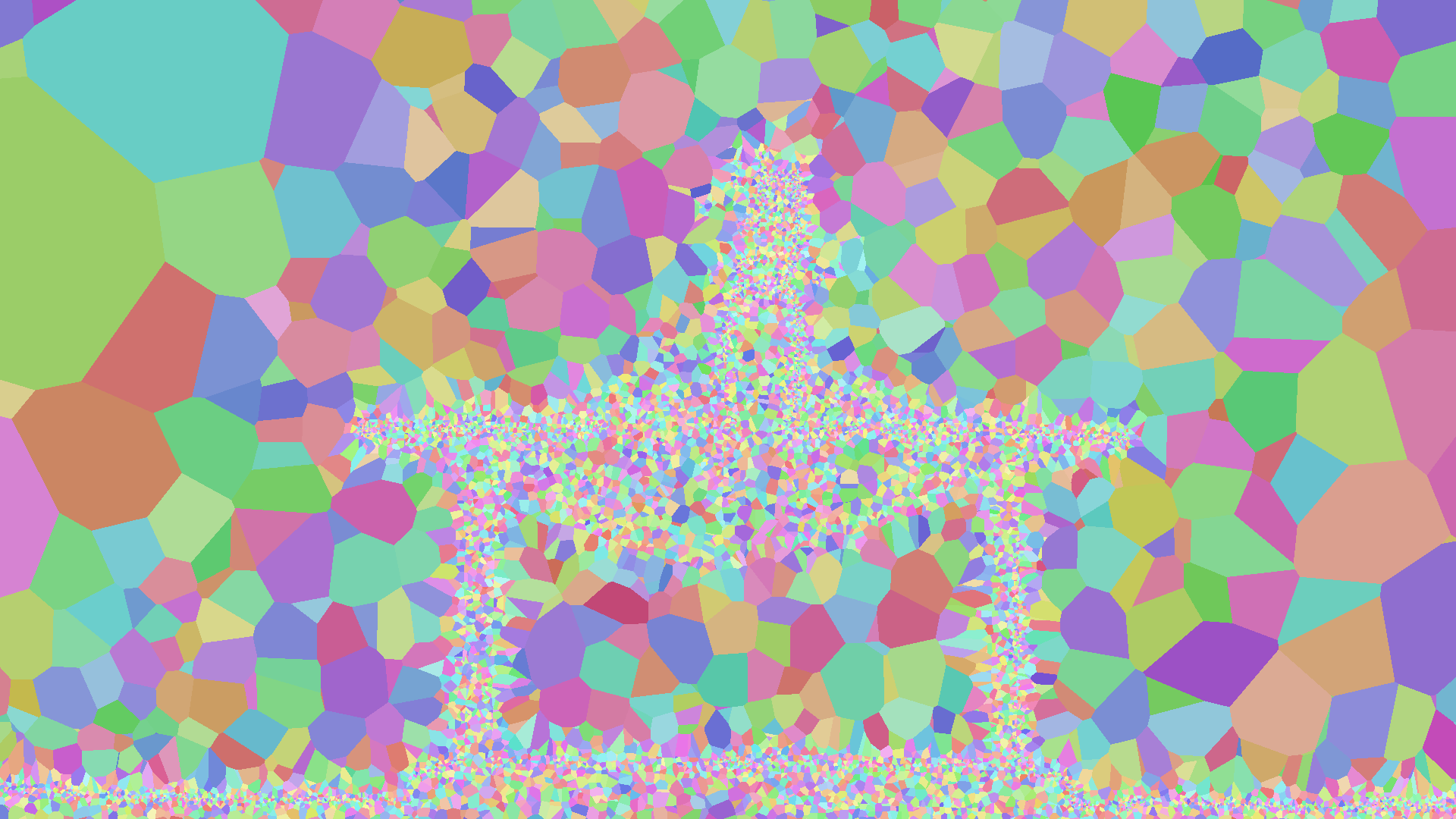}}
\captionof{figure}{
\textbf{Teaser --}
We introduce the Radiant Foam differentiable 3D representation, which can be used to learn accurate radiance fields for any novel view synthesis applications (left).
If we slice our foam along the plane highlighted by the red ``laser'', we expose (right) the internal structure of our representation: a polyhedral mesh that provides an injective parameterization of the 3D domain.
Our representation is a \textit{foam}, as the polyhedral cell structure is analogous to a closed-cell foam which partitions space into regions physically separated by thin, flat walls.
It is \textit{radiant}, as each foam bubble emits a view-dependent radiance that can be used to model the plenoptic function.
\vspace{1em}
}
\label{fig:teaser}
}]

\begin{abstract}
Research on differentiable scene representations is consistently moving towards more efficient, real-time models.
Recently, this has led to the popularization of splatting methods, which eschew the traditional ray-based rendering of radiance fields in favor of rasterization.
This has yielded a significant improvement in rendering speeds due to the efficiency of rasterization algorithms and hardware, but has come at a cost: the approximations that make rasterization efficient also make implementation of light transport phenomena like reflection and refraction much more difficult.
We propose a novel scene representation which avoids these approximations, but keeps the efficiency and reconstruction quality of splatting by leveraging a decades-old efficient volumetric mesh ray tracing algorithm which has been largely overlooked in recent computer vision research.
The resulting model, which we name \textit{Radiant Foam}, achieves rendering speed and quality comparable to Gaussian Splatting, without the constraints of rasterization.
Unlike ray traced Gaussian models that use \textit{hardware} ray tracing acceleration, our method requires no special hardware or APIs beyond the standard features of a programmable GPU.
\end{abstract}
\begin{figure}[b!]
\centering
\dropshadow{\includegraphics[height=.57\linewidth]{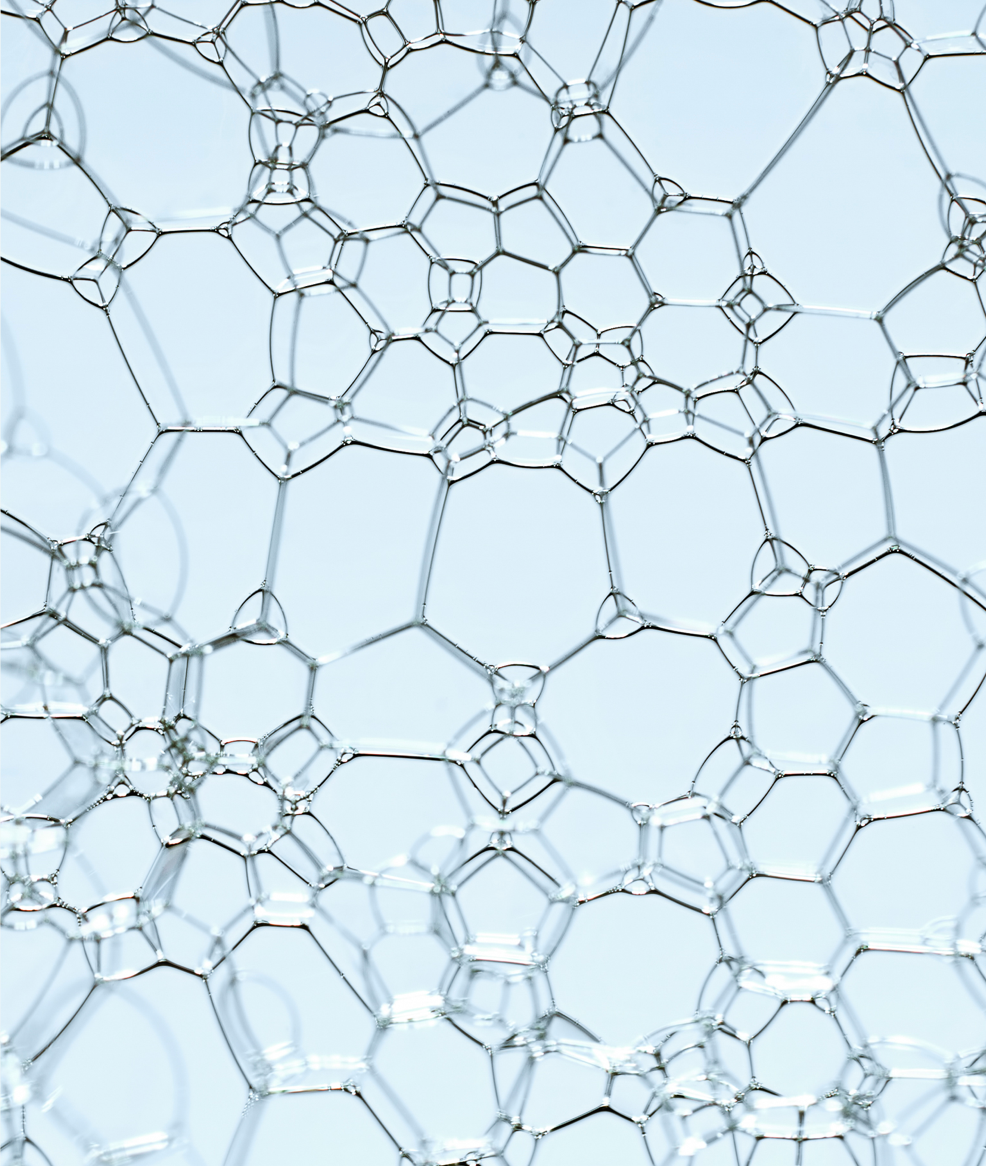}}
\hfill
\dropshadow{\includegraphics[height=.57\linewidth]{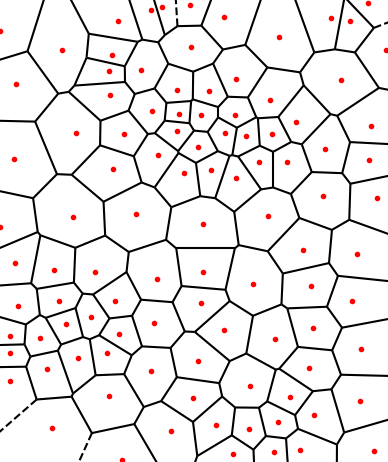}}
\caption{\textbf{Radiant Foam -- }
(left) In a stable foam, the pressure inside each bubble is roughly equal.
The interfaces between bubbles settle into shapes that balance forces, leading to polygonal cells resembling Voronoi patterns (right).
Our representation is nothing but a dense Voronoi tessellation of 3D space, where each point belongs to exactly one Voronoi cell.
The position of Voronoi sites is differentiable, making it amenable to gradient-based optimization.
}
\label{fig:subteaser}
\end{figure}

\begin{figure*}
\centering
\dropshadow{\includegraphics[width=0.32\textwidth]{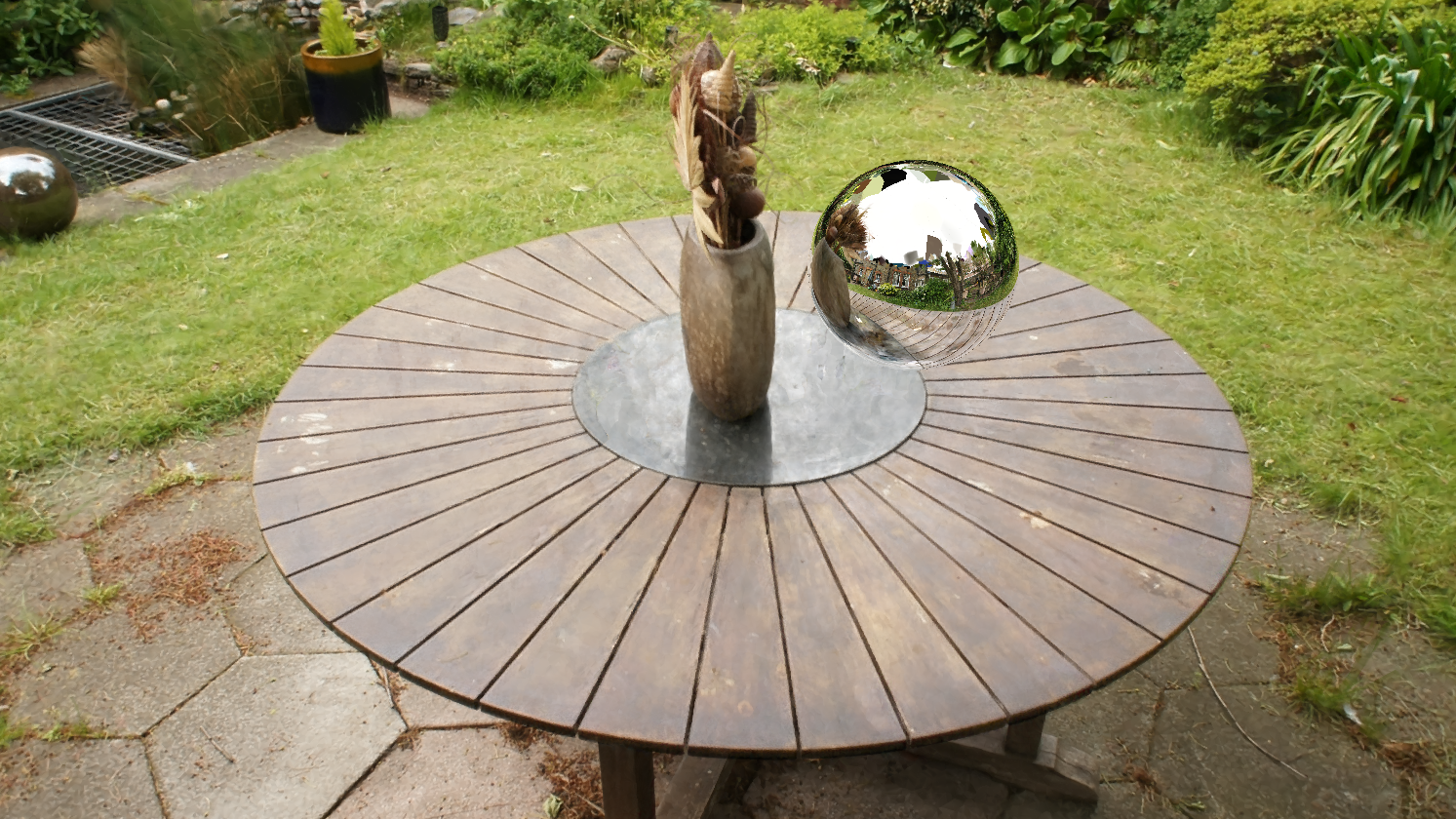}}
\hfill
\dropshadow{\includegraphics[width=0.32\textwidth]{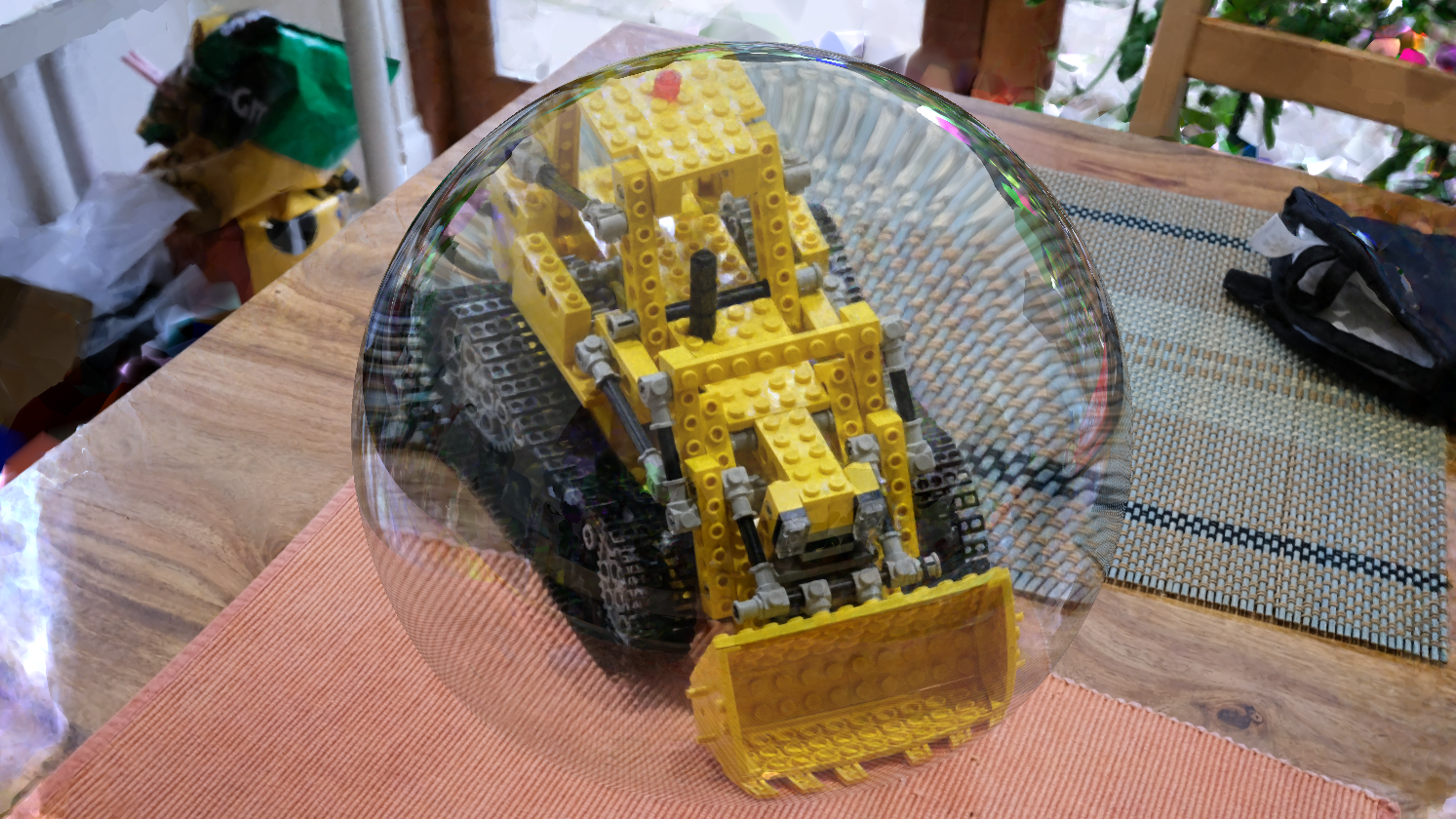}}
\hfill
\dropshadow{\includegraphics[width=0.32\textwidth]{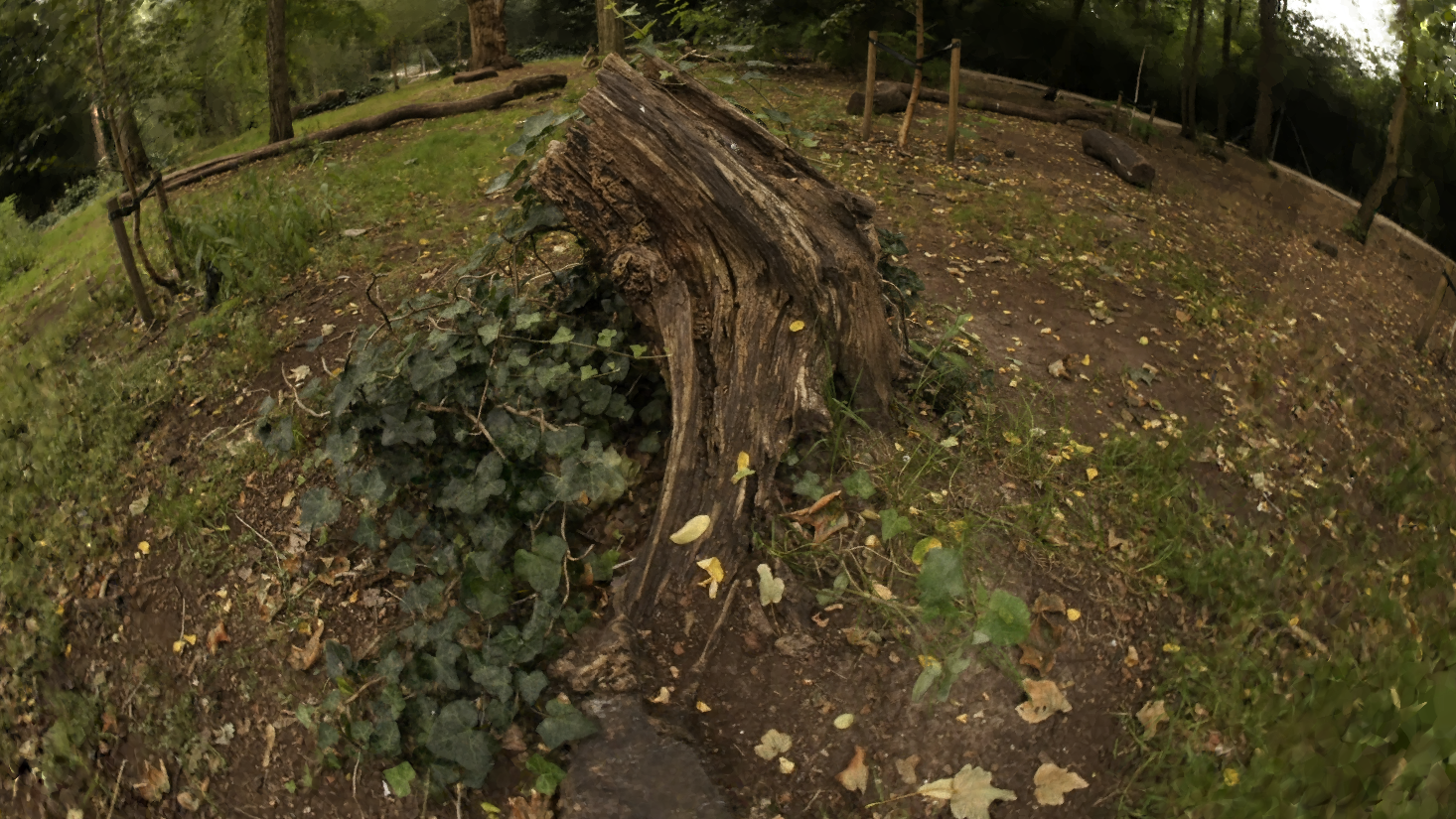}}
\caption{
{\bf Ray-based effects -- }
Ray tracing simplifies the implementation of many effects which are difficult to approximate with rasterization.
To motivate our work, we show here examples of integrating reflections, refractions, and non-linear camera models into our rendering pipeline.
Each would be complicated to achieve with rasterization, but requires only minor modification to our rendering code.
}
\label{fig:ray-based-effects}
\end{figure*}

\section{Introduction}
\label{sec:intro}
Neural radiance fields~(NeRF) have revolutionized 3D computer vision by allowing the extraction of dense 3D representations from collections of 2D images~\cite{nerf}.
In their early days, radiance fields suffered from low training and testing/rendering speed.
Since then, researchers have developed more efficient models~\cite{ingp}, as well as techniques for distilling models which enable highly efficient rendering/testing~\cite{mobilenerf}.
In particular, the real-time rendering performance was made possible by leveraging the rasterization pipeline readily available on GPU hardware.
Soon thereafter, researchers proposed to also incorporate rasterization into the training process, leading to the development of the now wildly popular 3D Gaussian Splatting~(3DGS)~\cite{gsplat}.

By comparing the development of radiance fields to that of classical graphics, one may argue that history is repeating itself: rasterization was introduced in computer graphics to \textit{approximate} the rendering equation~\cite{kajiya}, so to make it amenable to real-time rendering workloads.
However, light effects that are trivial to implement in ray tracing~(e.g. reflections,  refractions, and transparency; see \cref{fig:ray-based-effects}), are rather difficult to implement with rasterization.\footnote{After all, is this really that surprising, given that the physics of light is taught as the propagation of light rays through an environment?}
Consider the ``Graphics Gems'' series of books\footnote{\url{http://www.graphicsgems.org}} as a practical example, given how much of their content describes clever implementation tricks to express advanced light effects with rasterization engines.
Nonetheless, since the introduction of dedicated hardware support (NVIDIA RTX) in 2018, real-time rendering engines have, at least partially, returned to simulate light transport via ray tracing.

Recently, research on 3DGS representations has flourished, and much of this research is seemingly trying to fix issues \textit{caused} by rasterization, such as the removal of 3DGS popping artifacts~\cite{stopthepop}, or the introduction of more complex camera projection models~\cite{spherical3dgs}.
Others integrated ray tracing with 3DGS, so to accelerate rendering and enable more complex light behavior~\cite{3dgrt}.
But these developments have not discouraged researchers from investigating new, better, 3D representations.
The community craves a return to polygonal meshes, as meshes are the unquestionable workhorse of modern computer graphics.
And while we can find very interesting attempts at employing meshes for the modeling of radiance fields~\cite{tetrasplat, dmtet, tetranerf}, as we will later discuss, none of these has realistically been able to oust 3DGS as~\textit{the} representation for learning radiance fields.

Rather than revisiting history, and proposing clever engineering tricks that enable rasterization to work slightly better, we take a drastically different approach.
In particular, we highlight how, over two decades ago,~\citet{ray_plane} showed that fields represented by volumetric meshes admit a very efficient ray-tracing algorithm which requires no special hardware.
This approach has been subsequently overlooked in the resurgence of volume rendering methods, and we hope to re-introduce volumetric mesh models which can benefit from it to the differentiable rendering community.\footnote{It is also interesting to note that the volume rendering tutorial by~\citet{max05} that is cited in conjunction with~\citet{nerf} describes volumetric rendering within the context of volumetric meshes that \textit{partition} space, which is more similar to~\cite{fastray} than to~\cite{nerf}.}
In this paper, we make this representation differentiable, and carefully design refinement techniques for the underlying mesh.
These refinements allow us to accurately represent the surface of objects, and yet render efficiently by skipping empty portions of volume.

In a nutshell, our solution, which we name \texttt{Radiant Foam}~(\cref{fig:subteaser}) provides 3DGS-like rendering speed and quality, but has a training modality based on rays that resembles the one from NeRF~\cite{nerf}.
This implies that many NeRF techniques can now be seamlessly adapted to our method, with the significant advantage that the underlying geometry is \textit{explicitly} represented by a volumetric mesh.
We parameterize this volumetric mesh as a 3D Voronoi diagram~(\cref{fig:teaser}), which enables training of a mesh structure with dynamic connectivity through gradient descent by avoiding the discontinuities associated with discrete changes in other representations.
We also propose a coarse-to-fine training approach, which enables rapid construction of mesh models with adaptive resolution.

\begin{figure}
\centering
\includegraphics[width=\linewidth]{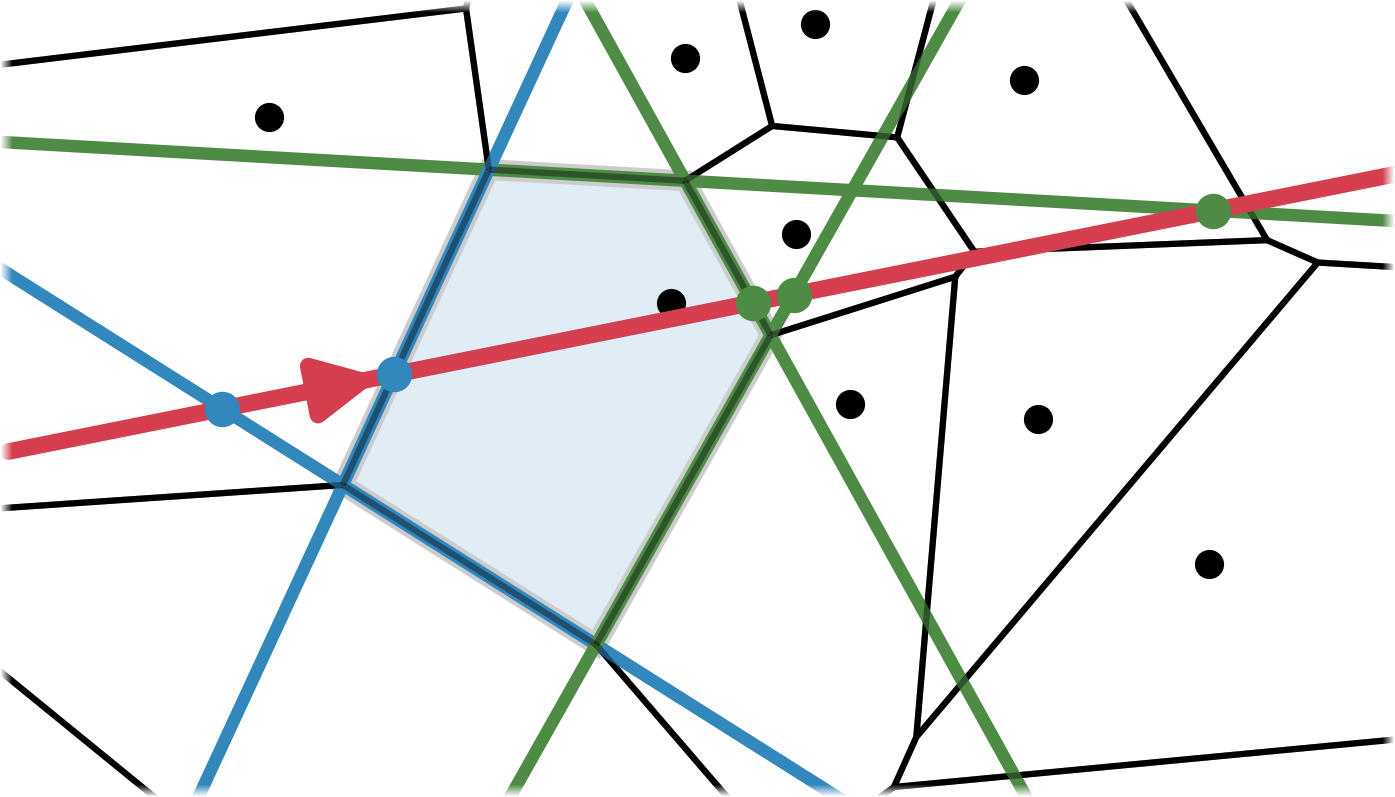}
\caption{
    {\bf Ray tracing foams -- }
    When a ray {\color{figred}(red)} enters a cell, we iterate through all the planar cell faces to identify the face through which the ray exits.
    This exit intersection is the first intersection along the ray with a normal vector less than 90 degrees from the ray direction~{\color{figgreen}(green)}; other intersections are considered back-facing~{\color{figblue}(blue)} and ignored.
    As the faces each correspond to a neighboring cell, the tracing then proceeds by stepping into the cell associated with the exit intersection and repeating the process.
}
\label{fig:ray_tracing}
\end{figure}
\section{Related Work}
\label{sec:related}
Neural Radiance Fields (NeRFs)~\cite{nerf} represent 3D scenes as volumetric radiance fields encoded within coordinate-based neural networks. 
This representation allows querying the network at any spatial location to retrieve volumetric density and view-dependent color, facilitating the generation of photo-realistic novel views.
The success of NeRFs has led to numerous follow-up works. 
For instance, significant effort has been devoted to enhancing training speed~\cite{ingp, kilonerf}, quality~\cite{mipnerf, mipnerf360}, and to extract surfaces from the representation~\cite{neus, volsdf}.
Finally, several works investigated ways to speed up the inference by baking the neural fields to more performant representations~\cite{mobilenerf, quadfields, smerf, adaptive-shells, binary-opacity-grids}.
While achieving high quality and fast rendering speeds, these methods often employ multi-stage training procedures.

\paragraph{Point-based rasterization}
Point-based rendering with primitives such as circles, spheres, or ellipsoids~\cite{surface-splatting, surfels, object-ewa} laid the foundations for point-based rasterization.
Differentiable rendering using depth-based blending has also been extended to volumetric particles in Pulsar~\cite{pulsar}, which employs sphere-based differentiable rasterization to render scenes with millions of spheres in real time.
While Pulsar provides a differentiable representation, \citet{gsplat} provides more expressive primitives in the form of soft, anistropic Gaussians, which can be differentiably optimized to fit the scene content.
This approach has inspired several follow-up works aimed at reducing its reliance on strong initialization~\cite{3dgs-mcmc}, reducing rendering time and memory footprints~\cite{light-3dgs, compressed-3dgs, reducing-mem-3dgs}, enabling the reconstruction of surfaces~\cite{sugar, 2dsplat}, and improving their ability to be trained for large spatial extents~\cite{octree-gs, hierarchicalgaussians}.

\paragraph{Extensions of 3DGS}
While significant progress has been made, these approaches still inherit the inherent limitations of rasterization. 
They struggle to handle camera distortions, model secondary lighting effects, or simulate sensor-specific properties like rolling shutter or motion blur. 
Recent efforts have aimed to address these challenges.
For example~\citet{radsplat} distilled a Zip-NeRF~\cite{zipnerf} into a 3DGS so to model distorted cameras and rolling shutter effects.
To capture secondary lighting effects, recent works have explored incorporating occlusion information into spherical harmonics for each Gaussian~\cite{relightable-3dgs, gs-ir}, or leveraging shading models and environment maps~\cite{gaussianshader}. 
These methods either rely on rasterization during inference~\cite{relightable-3dgs}, or during training~\cite{gs-ir, gaussianshader} hence inheriting the limitations of rasterization.
For complex lens effects, \citet{3dgs-on-the-move} modeled motion blur and rolling shutter by \textit{approximating} them in screen space through rasterization and pixel velocities.
\quad
Our approach introduces a principled framework for efficient ray tracing of volumetric primitives, overcoming the aforementioned limitations and enabling the simulation of effects like reflection, refraction, and camera distortion.

\begin{figure*}
\centering
\includegraphics[width=.32\linewidth]{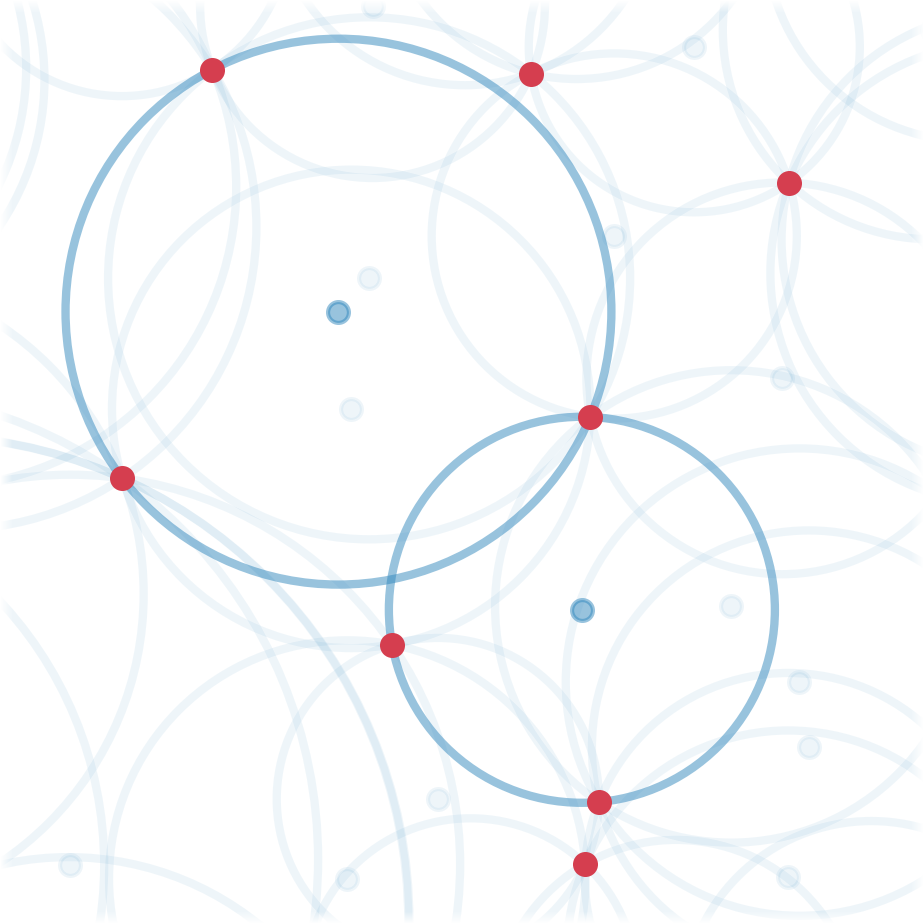}
\hfill
\includegraphics[width=.32\linewidth]{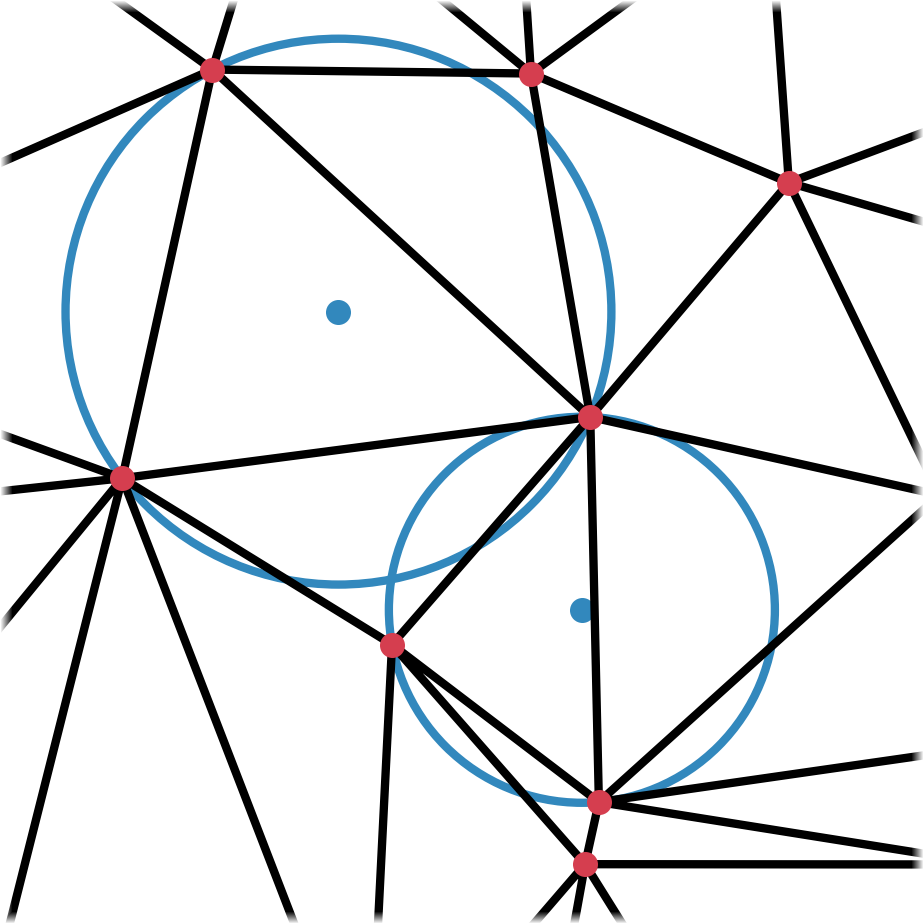}
\hfill
\includegraphics[width=.32\linewidth]{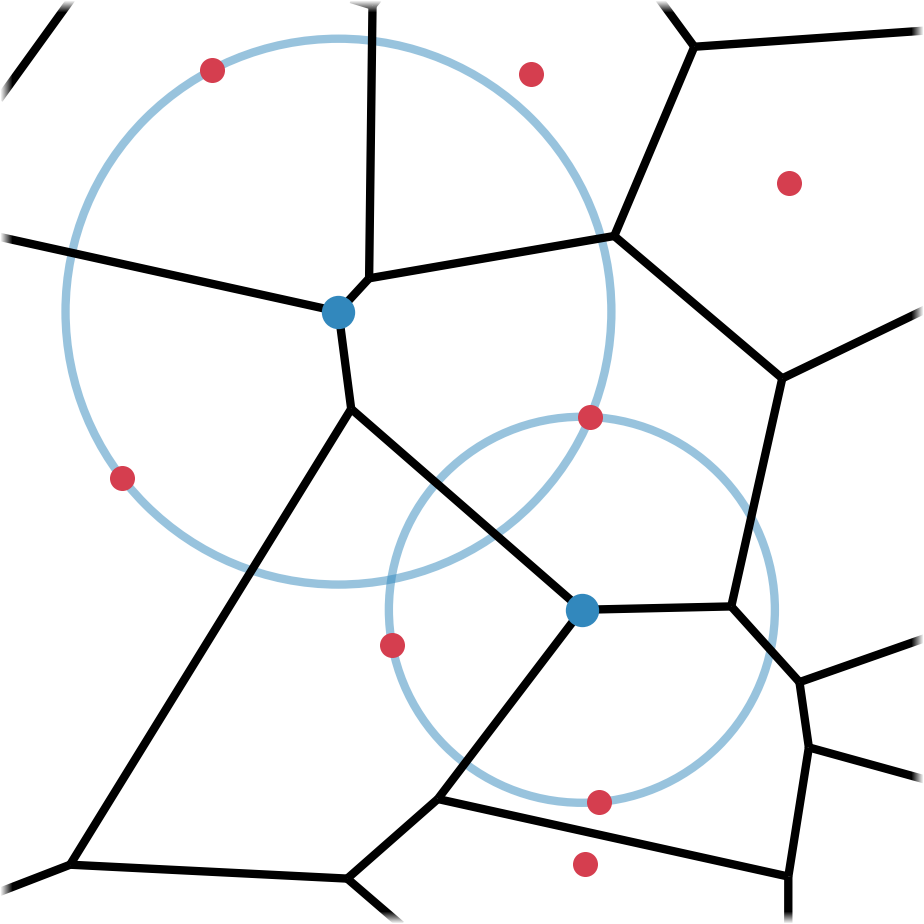}
\caption{
{\bf Delaunay and its dual Voronoi -- }
(left) Given a set of points {\color{figred}(red)} in $\real^N$, we can find circumspheres {\color{figblue}(blue)} which each pass through $N{+}1$ points.
(center) The set of all circumspheres which contain no points on their interior defines the Delaunay triangulation, where the $N{+}1$ points tangent to each circumsphere form a simplex.
(right) These circumspheres also describe the Delaunay triangulation's dual, the Voronoi diagram: the centers of circumspheres tangent to a point become the vertices of the Voronoi cell containing that point.
}
\label{fig:delaunay_voronoi}
\end{figure*}

\subsection{Ray tracing of volumetric primitives}
Ray tracing has been a cornerstone of photo-realistic rendering since its introduction~\cite{illumination-model}.
It enables the accurate simulation of light interactions with scene geometry, making it indispensable for applications requiring effects such as shadows, reflections, and refractions. 
Modern advancements in hardware have further accelerated ray tracing, allowing for real-time applications~\cite{harmonic-coordinates}.
Recently, \citet{3dgrt} proposed to use the NVIDIA OptiX ray tracer with 3DGS~\cite{gsplat} for fast ray-tracing.
While the method performs real-time ray tracing, it suffers from the tendency of 3DGS~\cite{gsplat} to produce overlapping primitives which degrade the quality of the hierarchical acceleration structure and increase the number of intersections per ray.
\quad
Our method represents the scene using a non-overlapping polyhedral mesh without the need for a secondary acceleration structure, thereby efficiently avoiding these limitations.

\subsection{Delaunay triangulation and Voronoi diagrams}
In computer graphics, Delaunay triangulations~\cite{delaunay} and Voronoi diagrams~\cite{voronoi} have been extensively studied for meshing (surfaces and volumes), and spatial partitioning~\cite{delaunay_voronoi_watson, voronoi_guibas, delaunay_shewchuk}.
Recently, they have also found novel applications in the realm of differentiable rendering.
DMTet~\cite{dmtet} introduced a deformable tetrahedral grid, generated using Delaunay triangulation, as an underlying 3D representation, while Tet-Splatting~\cite{tetrasplat} implements a differentiable rasterizer for this representation.
DeRF~\cite{derf} leverages Voronoi diagrams to spatially decompose a scene, resulting in faster training and inference, while Tetra-NeRF~\cite{tetranerf} employs a tetrahedral Delaunay mesh to model scenes more effectively.
While these methods inherit the ray-marching approach of~\cite{nerf} for volume rendering, they share a common limitation: slow rendering speeds caused by many MLP evaluations required per ray.
\quad
In contrast, with our method we propose to \textit{optimize} the mesh topology generated by the Voronoi diagram as a \textit{differentiable} polygonal mesh.
Our efficient ray tracing of volumetric particles also significantly accelerates the rendering speed.

\section{Method}
\label{sec:method}
Our method addresses the now-familiar problem of constructing representations of scenes from image collections.
As with NeRF~\cite{nerf} and its numerous successors, we achieve this reconstruction by gradient-based optimization of a differentiable scene representation.
In the following sections we propose a volumetric mesh-based differentiable representation, and explain how we are able to effectively optimize it from image supervision.

\begin{figure*}
\centering
\includegraphics[width=.31\linewidth]{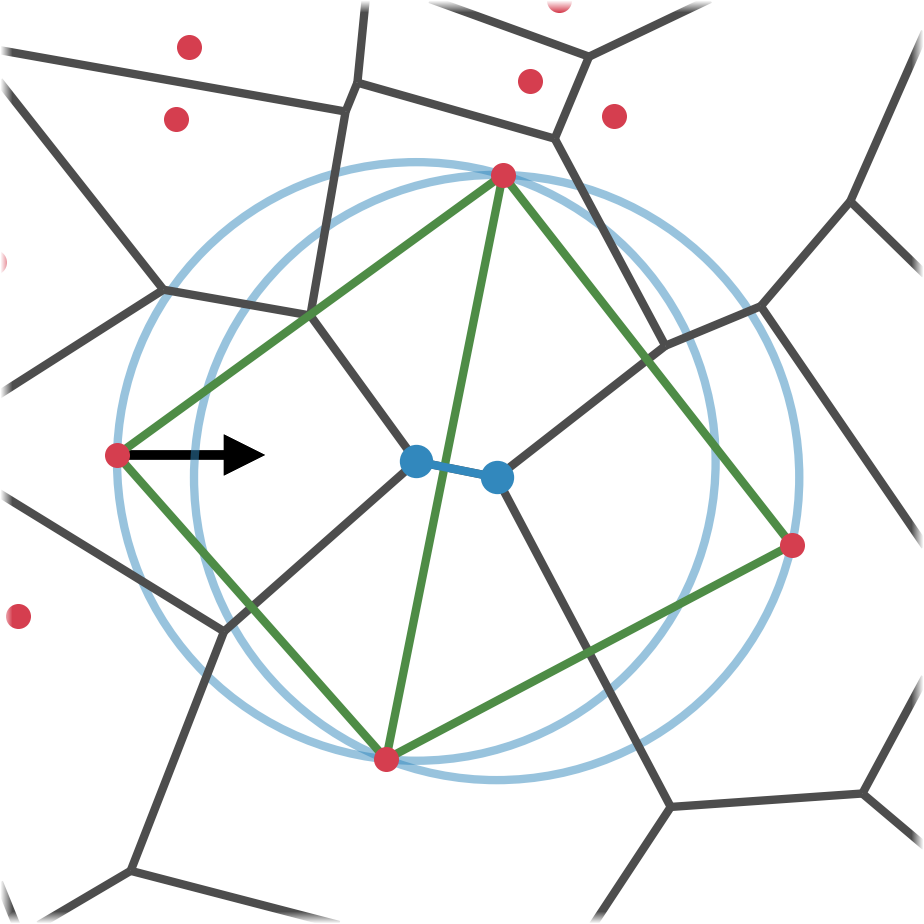}
\hfill
\includegraphics[width=.31\linewidth]{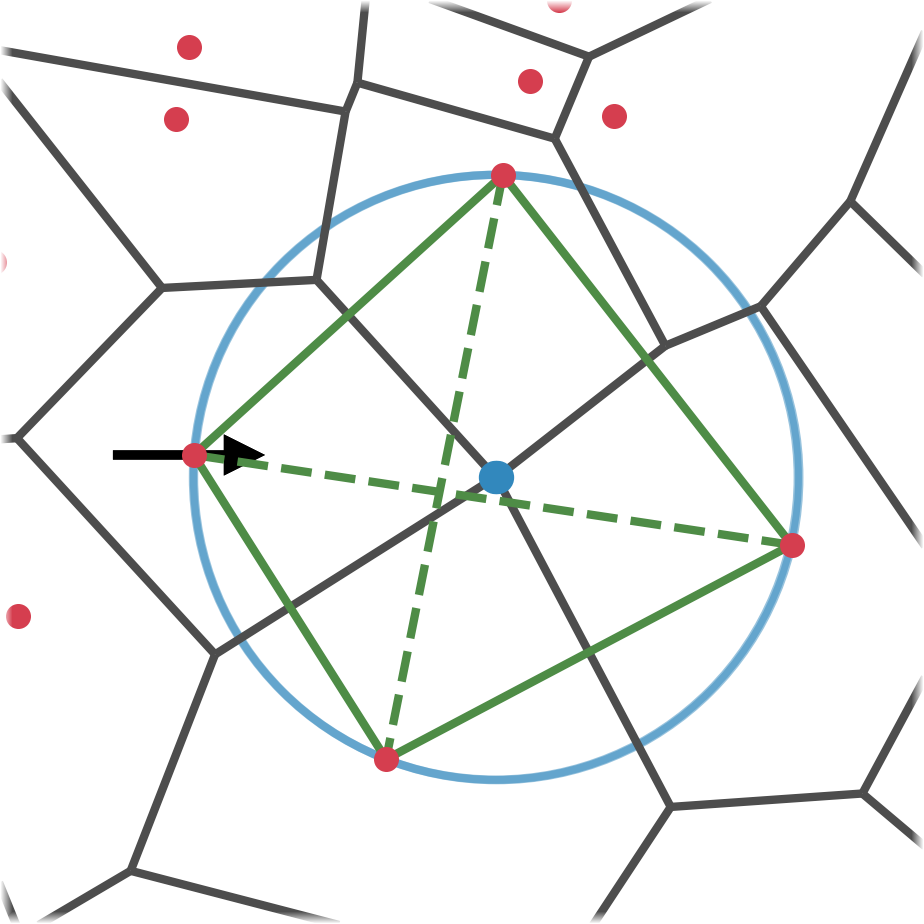}
\hfill
\includegraphics[width=.31\linewidth]{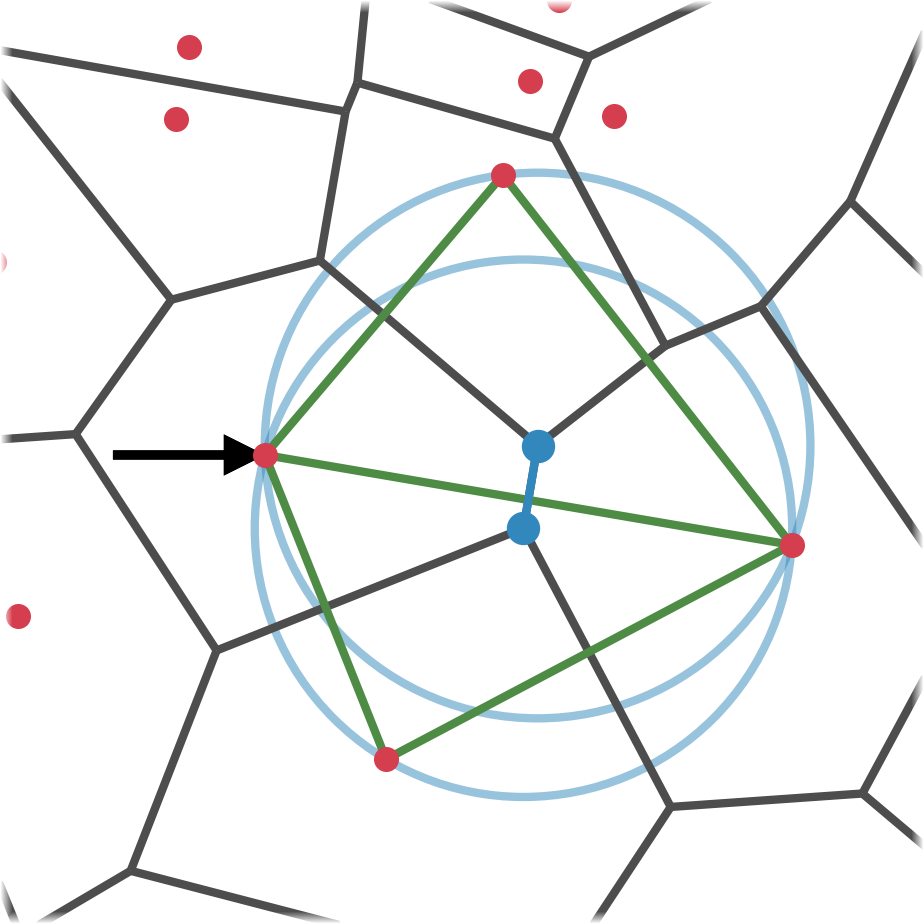}
\caption{
{\bf Edge flips -- }
The connectivity of the Delaunay graph {\color{figgreen}(green)} is sensitive to small positional perturbations, leading to ``edge flips'' in the triangulation.
These discrete changes occur at configurations where the circumspheres {\color{figblue}(blue)} of two neighbouring simplices become identical.
In the dual Voronoi diagram, this configuration also corresponds to a discrete change, but the cell boundary which changes has an area of zero at the moment of the flip (center).
Consequently, while there are still discrete changes in the adjacency structure of the Voronoi diagram, the \textit{shapes of the cells} vary continuously with the positions of the points {\color{figred}(red)}, which enables gradient-based optimization.}
\label{fig:edge_flip}
\end{figure*}
\subsection{Volume rendering}
Volume rendering has become the workhorse of modern differentiable scene reconstruction methods.
Volume rendering allows all points in space to make a continuously varying contribution to the observed color of viewing rays which pass through those points.
The effect of this contribution is controlled by a density field, which creates occlusions, and a radiance field, which determines the brightness of light observed.
Unlike many alternative rendering formulations, volume rendering of continuous fields is fully continuous with respect to all degrees of freedom, including both the viewpoint, as well as the density and radiance field values.
This property makes it very amenable to gradient-based optimization.
\quad
Volume rendering is defined by an integral over the segment $(t_\textrm{min}, t_\textrm{max})$ of a viewing ray.
Specifically the observed color $\radiance_\ray$ for a ray $\ray$ is:
\begin{align}
\radiance_\ray &= \int_{t_\textrm{min}}^{t_\textrm{max}} \transmittance(t) \cdot \density(\ray(t)) \cdot \radiance(\ray(t)) \; dt, \label{eq:volrend}\\
\transmittance(t) &= \textrm{exp}\left(-\int_{t_\textrm{min}}^t \density(\ray(u)) \; du \right),
\end{align}
where $\ray(t)$ denotes the point in 3D space at distance $t$ along ray $\ray$, and $\density(\cdot)$ and $\radiance(\cdot)$ denote the density and radiance fields respectively; see~\cite{digest} for more details.

\paragraph{Piecewise constant volumes}
In the case of \textit{piecewise constant} $\density(\cdot)$ and $\radiance(\cdot)$, the integral can be expressed as a sum over all $N$ ray segments with constant field values:
\begin{align}
\radiance_\ray &= \sum_{n=1}^{N} \transmittance_n \cdot (1 - \textrm{exp}(-\density_n \delta_n)) \cdot \radiance_n \: dt, \label{eq:volrendconst}\\
\transmittance_n &= \prod_{j=1}^n \textrm{exp}(-\density_j \delta_j),
\end{align}
where $\delta_n$ is the width of segment $n$.
This formulation admits a simple implementation as a loop over the segments in order of depth.
NeRF-based methods typically use \cref{eq:volrendconst} as an \textit{approximation} to \cref{eq:volrend}, with segments sampled according to some importance sampling scheme, e.g.~see~\cite{mipnerf360}.

Conversely, in our model these forms are \ul{exactly} equivalent.
We propose to leverage the algorithm by~\citet{ray_plane} along with a model of constant field values within the cells of a volumetric mesh~(see~\cref{fig:ray_tracing}), arriving at a representation for which \cref{eq:volrendconst} gives the \textit{exact} volume rendering result.
While this choice is highly advantageous in avoiding any complicated or expensive sampling schemes, we must take great care to not interfere with the continuity of the representation, which is critical for gradient-based optimizers.
In particular, the volumetric mesh itself receives gradients only through the segment widths $\delta_n$, which are determined by the locations of ray intersections with cell boundaries.
It is therefore critical that these boundary intersections \textit{vary continuously} with the optimizable parameters of the model.

\subsection{Differentiable mesh representation}

Constructing mesh representations of volumes or surfaces through gradient-based optimization is hardly a new problem, and many methods tackle it using a variety of strategies; see~\citet{dmtet} and citations therein.
Our method, however, encounters and must solve a challenge that most mesh optimization methods either avoid, or solve with discrete optimization techniques.
Generally, meshes are determined by degrees of freedom in two groups: vertex locations $\verts$ and cells $\cells$ (i.e. connectivity, or mesh topology).
Assuming all other parameters are fixed, optimizing for vertex locations is typically straightforward, as the intersections between rays and cell boundaries vary smoothly with vertex locations.
The true challenge arises if one desires to also 
optimize the connectivity $\cells$ of the representation.

\paragraph{The Delaunay triangulation}
Because mesh connectivity is inherently discrete, we can not optimize it directly with gradient-based methods.
To avoid this problem, we can define connectivity in terms of the vertex locations, such that each possible configuration of vertices corresponds to a \textit{unique} set of cells.
The most obvious choice for this mapping is the \emph{Delaunay triangulation}~\cite{delaunay}, which in 3D comprises the set of all tetrahedral cells $\cell_i {\in} \cells$ which may be formed from four vertices such that their circumspheres contain no additional vertices (the so-called ``Delaunay criterion'').
This construction is unique for point sets in general position (those lacking groups of 5 or more co-spherical points), and is easily computed using well-known and efficient algorithms; see \Cref{sec:implementation}.
This strategy encounters two significant problems if we wish to use it as a basis for differentiable volume rendering:
\begin{enumerate*} [label=(\arabic*)]
\item the discrete nature of the mesh connectivity is not entirely avoided, as the boundaries of Delaunay cells undergo discrete ``flips'' whenever a vertex enters the circumsphere of another cell.
These flips introduce \textit{discontinuities} into the optimization landscape, which interfere with the convergence of gradient descent; 
\item the number of tetrahedra in this model is not fixed in this model, and therefore it is not straightforward to associate each cell with optimizable $\density$ and $\radiance$ values as required for volume rendering.
\end{enumerate*}
The latter issue could be solved by associating field values with vertices (vs. tets), and interpolating those values within the cell, but this complicates volume rendering.

\paragraph{The Voronoi diagram}
Rather than utilize the Delaunay mesh directly and suffer its limitations, we instead look to the dual graph of the Delaunay triangulation: the \emph{Voronoi diagram}~\cite{voronoi}.
As shown in \Cref{fig:delaunay_voronoi}, this is constructed by placing a vertex at the circumcenter of each Delaunay tetrahedron; it partitions space into convex polyhedral cells $\cell_i{\in}\cells$ consisting of points which share a nearest neighbour among the primal vertices\footnote{Also known as Voronoi ``sites'' or ``seeds''.} $\primal_i$ of the Delaunay triangulation:
\begin{align}
\cell_i = \{x \in \real^3 : \argmin_j ||x - \primal_j|| = i\}.\label{eq:voronoi}
\end{align}
This may at first seem like a strange choice... after all, \textit{if the Delaunay Triangulation is unavoidably discontinuous in optimization, will not its dual also suffer this issue?}
\quad The answer to this question lies in the fact that any discrete change in the connectivity of the Voronoi Diagram exactly coincides with the point that the affected cell faces attain zero surface area.
As shown in \Cref{fig:edge_flip}, the discrete flips are effectively \textit{hidden} within these zero-volume regions of space, and the resulting field representation remains completely continuous with respect to the primal vertex locations.
\quad The number of cells in the Voronoi diagram is also constant regardless of configuration, which makes the association of $\density$ and $\radiance$ values to cells... trivial.
The resulting model is then effectively a \textit{learnable point cloud}, not dissimilar to the formulation of 3D Gaussian Splatting~\cite{gsplat}, though lacking the per-point covariance matrix.
\quad The only remaining issue is that our ray tracing algorithm~\cite{fastray} expects \emph{tetrahedral} cells.
We therefore modify the algorithm to handle the more general case of convex cells.
This modified tracing process pre-fetches the neighboring vertices of each vertex, thereby allowing an efficient iteration over the cell faces to find ray intersections.
For more detail on this algorithm, see~\Cref{fig:pseudocode}.

\begin{figure}[ht]
  \centering
  \begin{algorithm}[H]
    \caption{Ray tracing algorithm}\label{alg:render}
    \begin{algorithmic}[1]
      \Procedure{$\textsc{Render}$}{$\mathbf{o},\mathbf{d}$} \Comment{ray parameters}
        \State \(t_0 \gets 0\)
        \State \(i \gets \mathrm{nn}(\mathbf{o})\) \Comment{initial cell (nearest neighbour)}
        \State \(T \gets 1\)
        \State \(\mathbf{C} \gets \mathbf{0}\)
        \While{\(T > \epsilon\)}
          \State \(x \gets v_i\) \Comment{\(v_i\): primal vertex of cell \(i\)}
          \State \(t_1 \gets \infty\)
          \State \(i' \gets \varnothing\)
          \ForAll{\(j \in \mathrm{N}(i)\)}  \Comment{$\mathrm{N}(i)$: neighbours of cell \(i\)}
            \State \(x' \gets v_j\) \Comment{\(v_i\): primal vertex of cell \(j\)}
            \State \((t_j, \textrm{front}) \gets \textsc{Intersect}(\mathbf{o},\mathbf{d},x,x')\)
            \If{\(\textrm{front}~\textbf{and}~(t_j < t_1)\)}
              \State \(t_1 \gets t_j\)
              \State \(i' \gets j\)
            \EndIf
          \EndFor
          \State \(c \gets \mathbf{c}_{i}\) \Comment{$\mathbf{c}_{i}$: color of cell $i$}
          \State \(\sigma \gets \sigma_{i}\) \Comment{$\sigma_{i}$: density of cell $i$}
          \State \(\alpha \gets 1 - e^{-\sigma(t_1-t_0)}\)
          \State \(\mathbf{C} \gets \mathbf{C} + T\,\alpha\, c\)
          \State \(T \gets T\,(1-\alpha)\)
          \State \(t_0 \gets t_1\)
          \State \(i \gets i'\)
        \EndWhile
        \State \Return \(\mathbf{C}\)
      \EndProcedure
    \end{algorithmic}
  \end{algorithm}
\vspace{-1.5em}
  \caption{\textbf{Ray tracing --}
  Our ray tracing algorithm is simple, and based on the method proposed by \citet{ray_plane}.
  Unlike common algorithms for tracing triangle meshes and other unstructured scene representations, we do not require a hierarchical acceleration structure, and thus avoid the associated $O(\log(n))$ query operation.}
  \label{fig:pseudocode}
\end{figure}

\subsection{Optimization}
Similarly to~\citet{gsplat}, the (mostly) local nature of Voronoi cells renders the optimization landscape more prone to local minima.
We follow a similar strategy, by first carefully initializing the optimization, and then adaptively \textit{densifying} and \textit{pruning} Voronoi sites.
Additionally, to promote the formation of surface-like densities, we also employ a regularization objective similar to the \textit{distortion} loss~\cite{mipnerf360} commonly used by NeRF methods.

\paragraph{Densification}
Similarly to~\citet{gsplat}, to initialize training we start with a sparse point cloud obtained from~COLMAP~\cite{sfm}.
Over training, we perform densification and pruning operations to control the number of Voronoi sites and their density, allowing the model to adaptively re-allocate representational capacity to areas of space with more geometric and/or photometric detail.
\quad
We observe that gradients of the reconstruction loss with respect to Voronoi site locations can be used to identify cells which are underfitting the training signal.
We therefore use the norm of this gradient multiplied by the approximate radius of the cell as a measure of which cells require further densification.
Inspired by~\citet{3dgs-mcmc}, we select the candidates for densification by sampling a Multinomial distribution with probability mass function proportional to this measure.

\paragraph{Pruning}
Towards building a parsimonious representation, we remove cells from the Voronoi diagram that do not contribute to rendering.
However, it is not sufficient to simply delete ``empty'' cells (i.e. zero density), as the geometry of Voronoi cells is determined by the positions of adjacent sites, even when the density in those cells is zero.
To accurately represent object boundaries, it is therefore essential to retain cells with near-zero density that define the boundary. 
For this reason, our pruning strategy removes Voronoi sites that have very low density, \textit{and} are surrounded by neighbors with very low density.
This pruning ensures that we eliminate sites that neither contribute to nor define the surfaces, thereby maintaining the accuracy of the object boundaries.

\paragraph{Training objectives}
Similarly to the \textit{distortion} loss of Mip-NeRF 360~\cite{mipnerf360}, we apply a regularization on the distribution of contribution to the volume rendering integral along the ray.
This additional loss function encourages the density to concentrate at surfaces and reduces visible~``floater'' artifacts.
This objective is computed as
\begin{align}
\losst{quantile} = \expect_{t_1, t_2\sim\calU[0, 1]}[|W^{-1}(t_1) - W^{-1}(t_2)|],
\end{align}
where $W^{-1}(\cdot)$ denotes the quantile function (inverse CDF) of the volume rendering weight distribution along the ray.
This form has the same effect as the distortion loss~\cite{mipnerf360}, but avoids the need for a quadratic nested sum which would increase the computational cost and memory footprint of training.
Denoting with~$\losst{rgb}$ the typical L2 photometric reconstruction loss, our overall training objective is
\begin{equation}
\losst{} = \losst{rgb} + \lambda \losst{quantile}
\end{equation}

\subsection{Implementation details}
\label{sec:implementation}
We implemented our method in Python and C++ within the PyTorch framework, with custom CUDA kernels for ray tracing, Delaunay triangulation, and other operations requiring high efficiency.
This implementation includes an interactive viewer and a~(very) low overhead renderer, which we used to measure frame rates.
Importantly, \ul{nothing} in our implementation is dependent on 
dedicated ray tracing hardware, or the OptiX library, which is required by methods like~\cite{3dgrt}.
Therefore, with some engineering effort, our entire rendering loop could easily be implemented in a portable rendering framework like WebGL.

\paragraph{Training} 
Our training pipeline uses the Adam~\cite{adam} optimizer, and similarly to 3DGS, directly optimizes per-point position, density and view-dependent color via spherical harmonics of degree three.
We use the softplus activation function with $\beta{=}10$ for density to constraint it within the~$[0, \infty)$ range, while keeping smooth gradients.
For the location of points, we start at a learning rate of $2e^{-4}$ and decay it using a cosine annealing scheduler to a final learning rate of $2e^{-6}$.
For point density and spherical harmonics, we start at a learning rate of $1e^{-1}$ and $5e^{-3}$ respectively and decay it with a cosine annealing scheduler by a factor of $0.1$.
Following 3DGS~\cite{gsplat}, we optimize only the zero-order component of SH coefficients and the high-order coefficients with a warmup for the first 25\% of the total training iterations.
\quad
Similarly to \cite{revising-densification, 3dgs-mcmc}, after initialization and warm-up training, we gradually grow the number of Voronoi sites so that points are placed at useful locations.
We progressively increase the number of points up to half the total training iterations, linearly increasing the number of points until the maximum desired number of points is obtained.

\paragraph{Voronoi optimization}
We maintain an adjacency data structure throughout training, which defines the Voronoi cells for rendering.
Whenever the primal vertex positions are changed we must update the adjacency by performing an incremental Delaunay triangulation.
While much faster than a complete rebuild, the incremental update is still computationally expensive for large point sets, so we allow the optimizer to take multiple steps between mesh rebuilds.
We start at a 1:1 ratio after each densification and increase to 1:100, as the frequency of discrete changes to the mesh decreases over time with the converging optimization.
This strategy balances the overall speed of training with maintaining a relatively accurate mesh structure.
All our experiments are optimized for 20k iterations, with the last 2k only updating radiance and density attributes while positions are frozen.
This process takes, as an example, 70 minutes on the ``bonsai'' scene with an RTX 4090 GPU.

\begin{table}
\centering
\resizebox{\linewidth}{!}{
\setlength{\tabcolsep}{2pt}
\begin{tabular}{l|cccr|cccr}
    \toprule
    & \multicolumn{4}{c|}{\mipnerf} & \multicolumn{4}{c}{\deepblend} \\
    & PSNR$\uparrow$ & SSIM$\uparrow$ & LPIPS$\downarrow$ & FPS$\uparrow$ & PSNR$\uparrow$ & SSIM$\uparrow$ & LPIPS$\downarrow$ & FPS$\uparrow$ \\
    \midrule 
    \multicolumn{1}{c}{} & \multicolumn{8}{c}{Rasterization} \\
    \midrule 
    3DGS*~\cite{gsplat} & 28.69 & 0.87 & 0.22 & 293 & 29.41 & \textbf{0.90} & \textbf{0.32} & 319 \\
    Mip-Splatting~\cite{mip-splatting} & 29.39 & 0.88 & 0.20 & 241 & 29.47 & \textbf{0.90} & \textbf{0.32} & 260 \\
    3DGS-MCMC~\cite{3dgs-mcmc} & \textbf{29.72} & \textbf{0.89} & \textbf{0.19} & \textbf{302} & \textbf{29.71} & \textbf{0.90} & \textbf{0.32} & \textbf{662} \\
    \midrule
    \multicolumn{1}{c}{} & \multicolumn{8}{c}{Ray Tracing / Marching} \\
    \midrule 
    Plenoxels~\cite{plenoxels} & 23.63 & 0.67 & 0.44 & {$<30$} & 23.06 & 0.80 & 0.51 & {$<30$} \\
    iNGP-Big~\cite{ingp} & 26.75 & 0.75 & 0.30 & {$<30$} & 24.96 & 0.82 & 0.39 & {$<30$} \\
    MipNeRF360~\cite{mipnerf360} & \textbf{29.23} & \textbf{0.84} & \textbf{0.21} & {$<$}1 & \textbf{29.40} & \textbf{0.90} & \textbf{0.25} & {$<$}1 \\
    3DGRT**~\cite{3dgrt} & 28.71 & 0.85 & 0.25 & 78 & 29.23 & \textbf{0.90} & 0.32 & 119 \\
    \textbf{RadiantFoam~(v1)} & 28.47 & 0.83 & \textbf{0.21} & \textbf{200} & 28.95 & 0.89 & 0.26 & \textbf{301} \\
    \bottomrule
\end{tabular}
}
\caption{
    {\bf Novel view synthesis --}
    We evaluate our method's accuracy in reconstructing held-out views for two standard datasets.
    Our method has similar performance to 3DGS and 3DGSRT, while providing \textit{significantly} higher frame rates than the latter.
    FPS is measured as an average on the test set for each scene.
    *For 3DGS, we report corrected LPIPS scores provided by \citet{revising-densification}
    **Note that 3DGRT FPS results were measured with an RTX 6000 Ada GPU.
    We report the results from the original publication, as the code is not open source.
}
\label{tab:qual_res}
\end{table}

\section{Experiments}
\label{sec:experiments}
%
We evaluated our algorithm on a total of 9 real-world scenes sourced from two publicly available datasets. 
Specifically, we utilized the complete set of scenes from the Mip-NeRF 360 dataset~\cite{mipnerf360} except for two private scenes (flowers and treehill) and two scenes from the Deep Blending dataset~\cite{deepblending}.
These datasets contain scenes with a diverse range of capture styles, including bounded indoor scenes and expansive, unbounded outdoor environments.
For Mip-NeRF 360, to make our results compatible with~\cite{gsplat, 3dgrt}, we downsample images for the indoor scenes by a factor of two, and the outdoor scenes by four.
For Deep Blending scenes, we use the original image resolutions.
All frame rates were measured on a consumer-grade RTX 4090 GPU.

\paragraph{Metrics}
We assess each method using three widely recognized image quality metrics: Peak Signal-to-Noise Ratio (PSNR), Structural Similarity Index (SSIM), and Learned Perceptual Image Patch Similarity (LPIPS).
In the supplementary web page we include rendered video paths for selected scenes, showcasing views significantly different from the input images.

\paragraph{Quantitative results}
We report our quantitative results for the Mip-NeRF 360~\cite{mipnerf360} and Deep Blending~\cite{deepblending} in \Cref{tab:qual_res}.
In terms of quality, our method achieves results comparable to, or slightly below, those of 3DGS~\cite{gsplat} and 3DGRT~\cite{3dgrt}~(the state-of-the-art differentiable ray tracing method).
However, our method \ul{excels} in rendering speed. 
As shown in \Cref{tab:qual_res}, our efficient ray-tracing implementation achieves in some cases over 300 FPS, more than twice as fast as 3DGRT~($119$ FPS), while maintaining a similar rendering speed to rasterization methods.

\begin{table}[!tb]
\begin{center}
\resizebox{\linewidth}{!}{
    \setlength{\tabcolsep}{3pt}
    \begin{tabular}{cccc|c|c|c|c}
    \toprule
     SfM & Densify & Prune & Quantile & Bonsai & Garden & Playground & Mean \\
    \midrule
    \xmark & \cmark & \cmark & \cmark & 29.65 & 25.83 & 26.34 & 27.00 \\
    \xmark & \xmark & \cmark & \cmark & 20.23 & 18.88 & 19.55 & 19.36 \\
    \cmark & \cmark & \xmark & \cmark & \textbf{32.25} & \textbf{26.58} & 29.46 & \textbf{29.15} \\
    \cmark & \cmark & \cmark & \xmark & 29.62 & 25.35 & \textbf{29.59} & 27.90 \\
    \cmark & \cmark & \cmark & \cmark & 32.15 & \textbf{26.58} & \textbf{29.59} & \textbf{29.15} \\
    \bottomrule
    \end{tabular}
}
\end{center}
\vspace{-1em}
\caption{{\bf Ablation table --}  We evaluate the impact of various components in our method by systematically excluding them and analyzing the reconstruction quality (PSNR$\uparrow$) on the Bonsai and Garden scenes from MipNeRF 360~\cite{mipnerf360}, as well as the Playground scene from Deep Blending~\cite{deepblending}.}
\label{tab:ablation}
\end{table}
\subsection{Ablation}
\begin{figure*}
    \centering
    \begin{overpic}[width=\linewidth]{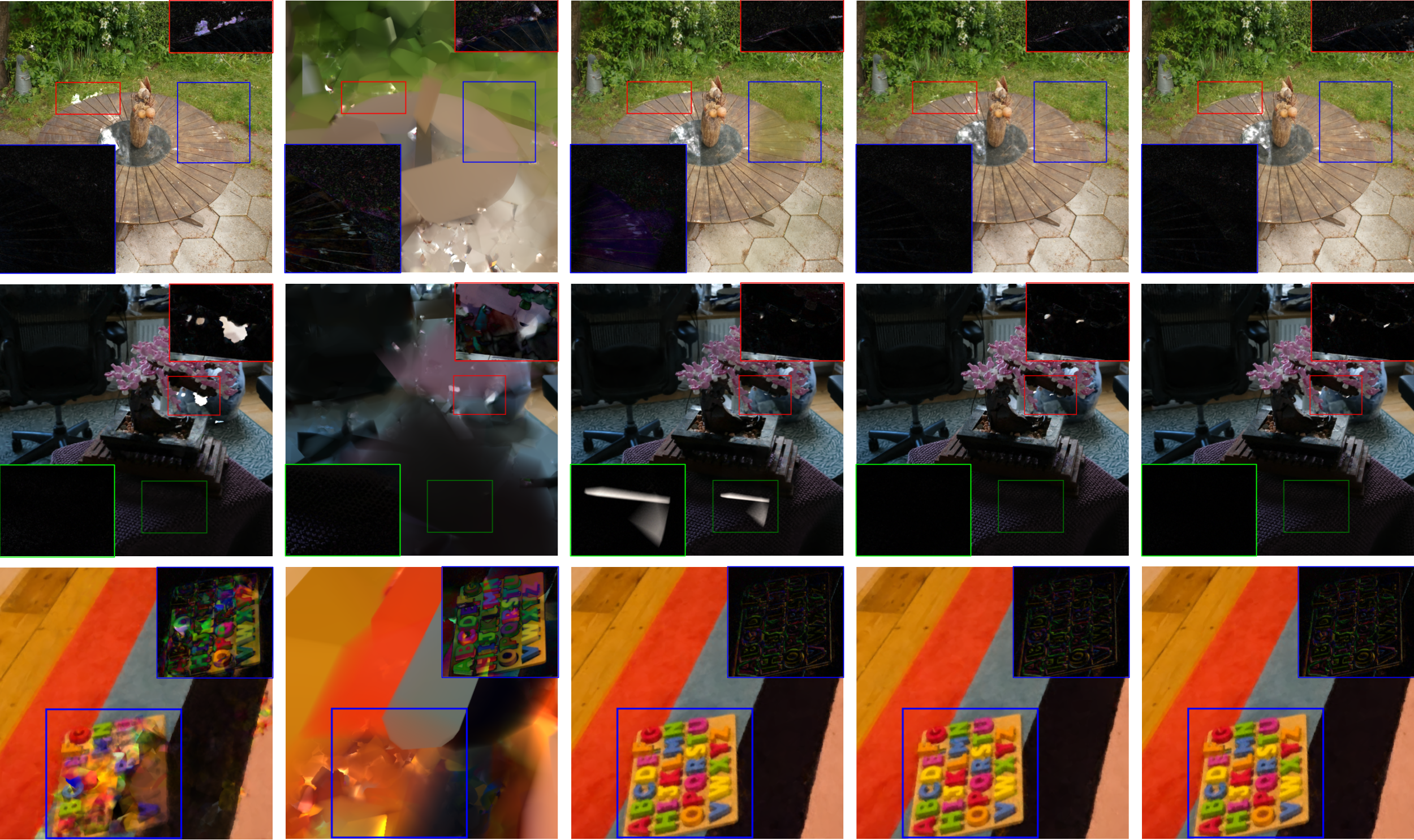}
    \put(6, -3){{No SFM}}
    \put(23, -3){{No Densify+SFM}}
    \put(46, -3){{No Quantile}}
    \put(66, -3){{No Prune}}
    \put(89, -3){{Full}} 
    \end{overpic}
    \vspace{0.5pt}
    \caption{
        {\bf Ablations -- } 
        We qualitatively analyze the impact of excluding various components from our method. 
        For better visualization, we zoom in on specific image regions and present their corresponding error maps.
        \textit{No SFM}: While the method performs relatively well without initialization, it degrades in sparsely covered regions of the training views. 
        \textit{No Densify+SFM}: Without densification, the method significantly underperforms, leading to under-represented scenes. 
        \textit{No Quantile}: Disabling the quantile loss introduces floaters, degrading rendering quality. 
        \textit{No Pruning}: Ceasing point pruning has minimal impact due to the low number of prunable points.
    }
    \label{fig:ablation}
\end{figure*}
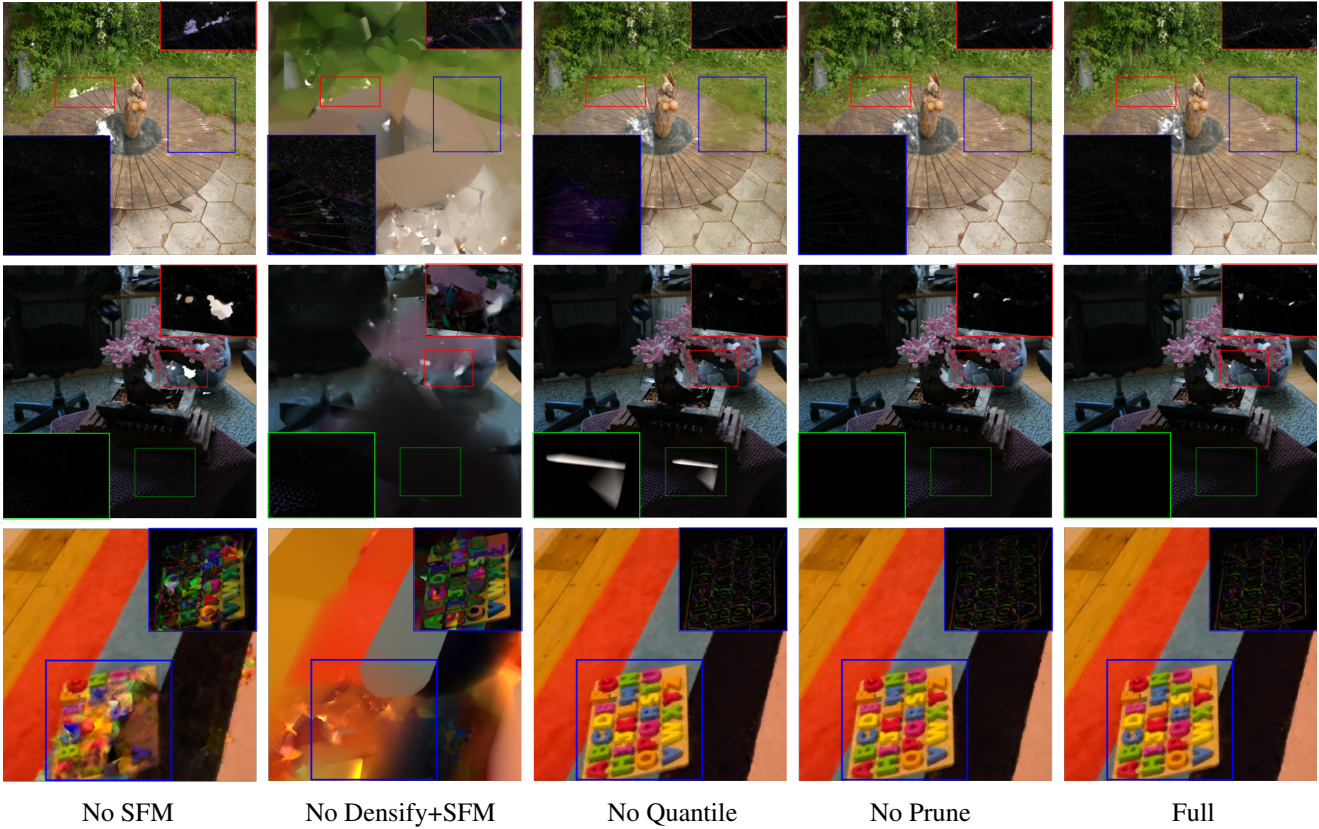
We isolated the different contributions and choices we made and constructed a set of experiments to measure their effect. 
Specifically, we test the following aspects of our algorithm: initialization from SfM, our densification and pruning strategies, and regularization loss. 
The quantitative effect of each choice is summarized in \Cref{tab:ablation}.

\paragraph{Initialization from SFM} 
We assess the importance of initializing our Voronoi sites from the SfM point cloud. 
For this, we initialize our representation with $2^{17}$ points from a normal distribution with a standard deviation of 10.
We observe that our method performs relatively well even without the SfM points. 
Instead, it degrades mainly in the background and tends to have more floaters in regions that are sparsely covered in the training views, see \Cref{fig:ablation}.

\paragraph{Densification and pruning}
Due to numerical approximations, our triangulation algorithm can fail when processing very close (or identical) points. 
This limitation prevents us from initializing our representation directly with the final set of points obtained by duplicating and perturbing the Structure from Motion (SfM) points, as SfM typically produces numerous closely spaced points.
Consequently, for the ablation study on densification, we utilize the random initialization strategy described earlier to initialize our representation with the intended number of points.
We observe that our method significantly underperforms without densification, resulting in an under-represented scene where resources are not adequately allocated to regions with complex geometry or texture.
\quad
In the ablation study of our pruning strategy, we cease pruning those Voronoi sites that are neither surface nor boundary points. 
We notice that this pruning approach has minimal impact on the quality of the renderings because the number of prunable points is very low. 
By starting with a sparse point set and progressively densifying only in under-represented areas, we ensure that points are allocated exclusively to necessary regions, resulting in few prunable points.

\paragraph{Quantile loss}
We disable the quantile regularization while training, which results in floater artifacts that degrade rendering quality for novel views.
Some scenes are more affected by this than others, as the floaters result from data-dependent ambiguities, but on average we find that the quantile regularization improves reconstruction metrics.

\section{Conclusions}
We have introduced Radiant Foam, a novel representation that allows real-time differentiable ray tracing.
The core of our method is a foam structure of polyhedral cells, which allows efficient volumetric mesh ray tracing algorithms to be applied \textit{without} relying on dedicated hardware such as NVIDIA RT cores.
We allow these cells to be continuously optimized by parameterizing them as a Voronoi diagram, which we show to be differentiable under volume rendering.
By doing so, we have shown that one can achieve similar modeling quality as 3D Gaussian Splatting, but without sacrificing the benefits of a true ray tracing-based volume renderer, nor the fast rendering speed of rasterization-based renderers.

\paragraph{Limitations and future work}
While the Voronoi-based representation we have proposed is very effective at constructing foam models through continuous optimization, the space of possible foam models which could be used in our rendering pipeline is much larger than what is parameterized by Voronoi.
Most notably, our current model always requires that cell boundaries be equidistant between neighbouring points, which leads to many small, empty cells being needed to define a surface.
Future work could potentially relax this requirement by generalizing beyond Voronoi diagrams.
\quad
Other open research questions include how to compose multiple foam models together efficiently and accounting for varying illumination, how to model dynamic content instead of static scenes, how to enable editing of scenes, and how to integrate generative modeling with our representation.
Progress in these directions could make foam models relevant in real-time ray tracing applications currently dominated by triangle meshes, as we have already found that foam-based ray tracing can exceed the performance of dedicated ray tracing hardware.

\paragraph{Acknowledgements}
This work was supported in part by the Natural Sciences and Engineering Research Council of Canada (NSERC) Discovery Grant [2023-05617], NSERC Collaborative Research and Development Grant, the SFU Visual Computing Research Chair, Google Research, Digital Research Alliance of Canada, and Advanced Research Computing at the University of British Columbia.
\quad
We would like to thank George Kopanas, Lily Goli, Alex Evans, Thomas Muller, Bernhard Kerbl, Vincent Sitzmann, Forrester Cole, Or Litany, and David Fleet for their feedback and/or early research discussions.


\vfill

\section{Per Scene metrics}
\begin{table*}
\centering
\resizebox{\linewidth}{!}{
\setlength{\tabcolsep}{5pt}
\begin{tabular}{l|ccc|cccc}
    \toprule
    & 3DGS~\cite{gsplat} & Mip-Splatting~\cite{mip-splatting} & 3DGS-MCMC~\cite{3dgs-mcmc} & Plenoxels~\cite{plenoxels} & iNGP-Big~\cite{ingp} & MipNerf360~\cite{mipnerf360} & Ours \\
    & \scriptsize{PSNR$\uparrow$ / SSIM$\uparrow$ / LPIPS$\downarrow$} & \scriptsize{PSNR$\uparrow$ / SSIM$\uparrow$ / LPIPS$\downarrow$} & \scriptsize{PSNR$\uparrow$ / SSIM$\uparrow$ / LPIPS$\downarrow$} & \scriptsize{PSNR$\uparrow$ / SSIM$\uparrow$ / LPIPS$\downarrow$} & \scriptsize{PSNR$\uparrow$ / SSIM$\uparrow$ / LPIPS$\downarrow$} & \scriptsize{PSNR$\uparrow$ / SSIM$\uparrow$ / LPIPS$\downarrow$} & \scriptsize{PSNR$\uparrow$ / SSIM$\uparrow$ / LPIPS$\downarrow$} \\
    \midrule
    Room & 30.63 / 0.91 / 0.27 & 31.74 / \textbf{0.93} / 0.27 & \textbf{32.30} / \textbf{0.93} / \textbf{0.25} & 27.59 / 0.84 / 0.42 & 29.69 / 0.87 / 0.26 & \textbf{31.63} / \textbf{0.91} / 0.21 & 30.87 / \textbf{0.91} / \textbf{0.19} \\
    Counter & 28.70 / 0.91 / 0.24 & \textbf{29.16} / \textbf{0.92} / 0.24 & \textbf{29.16} / 0.91 / \textbf{0.23} & 23.63 / 0.76 / 0.44 & 26.69 / 0.82 / 0.31 & \textbf{29.55} / \textbf{0.89} / 0.20 & 28.59 / 0.88 / \textbf{0.19} \\
    Bonsai & 31.98 / 0.94 / 0.24 & 32.31 / \textbf{0.95} / \textbf{0.23} & \textbf{32.67} / \textbf{0.95} / \textbf{0.23} & 24.67 / 0.81 / 0.40 & 30.69 / 0.91 / 0.21 & \textbf{33.46} / \textbf{0.94} / 0.18 & 32.15 / 0.93 / \textbf{0.17} \\
    Kitchen & 30.32 / 0.92 / \textbf{0.14} & 31.55 / 0.93 / 0.15 & \textbf{32.23} / \textbf{0.94} / \textbf{0.14} & 23.42 / 0.65 / 0.45 & 29.48 / 0.86 / 0.20 & \textbf{32.23} / \textbf{0.92} / \textbf{0.13} & 31.40 / 0.91 / \textbf{0.13} \\
    Bicycle & 25.25 / 0.77 / 0.23 & 25.72 / \textbf{0.78} / \textbf{0.19} & \textbf{26.06} / \textbf{0.78} / \textbf{0.19} & 21.92 / 0.50 / 0.51 & 22.17 / 0.51 / 0.45 & \textbf{24.37} / \textbf{0.69} / \textbf{0.30} & 24.19 / 0.68 / 0.31 \\
    Garden & 27.41 / 0.87 / 0.12 & 27.76 / \textbf{0.88} / \textbf{0.11} & \textbf{27.99} / 0.87 / \textbf{0.11} & 23.49 / 0.61 / 0.39 & 25.07 / 0.70 / 0.26 & \textbf{26.98} / 0.81 / 0.17 & 26.58 / \textbf{0.82} / \textbf{0.16} \\
    Stump & 26.55 / 0.78 / 0.24 & 26.94 / \textbf{0.79} / 0.21 & \textbf{27.67} / 0.78 / \textbf{0.20} & 20.66 / 0.52 / 0.50 & 23.47 / 0.59 / 0.42 & \textbf{26.40} / \textbf{0.74} / \textbf{0.26} & 25.48 / 0.71 / 0.29 \\
    \bottomrule
\end{tabular}
}
\caption{
   PSNR, SSIM, and LPIPS scores for Mip-NeRF360~\cite{mipnerf360} scenes.
    }
\label{tab:mipnerf360_all}
\end{table*}

\begin{table*}
\centering
\resizebox{\linewidth}{!}{
\setlength{\tabcolsep}{5pt}
\begin{tabular}{l|ccc|cccc}
    \toprule
    & 3DGS~\cite{gsplat} & Mip-Splatting~\cite{mip-splatting} & 3DGS-MCMC~\cite{3dgs-mcmc} & Plenoxels~\cite{plenoxels} & iNGP-Big~\cite{ingp} & MipNerf360~\cite{mipnerf360} & Ours \\
    & \scriptsize{PSNR$\uparrow$ / SSIM$\uparrow$ / LPIPS$\downarrow$} & \scriptsize{PSNR$\uparrow$ / SSIM$\uparrow$ / LPIPS$\downarrow$} & \scriptsize{PSNR$\uparrow$ / SSIM$\uparrow$ / LPIPS$\downarrow$} & \scriptsize{PSNR$\uparrow$ / SSIM$\uparrow$ / LPIPS$\downarrow$} & \scriptsize{PSNR$\uparrow$ / SSIM$\uparrow$ / LPIPS$\downarrow$} & \scriptsize{PSNR$\uparrow$ / SSIM$\uparrow$ / LPIPS$\downarrow$} & \scriptsize{PSNR$\uparrow$ / SSIM$\uparrow$ / LPIPS$\downarrow$} \\
    \midrule
    Dr Johnson & 28.77 / \textbf{0.90} / 0.33 & 28.76 / \textbf{0.90} / \textbf{0.32} & \textbf{29.05} / 0.89 / 0.33 & 23.14 / 0.79 / 0.52 & 28.26 / 0.85 / 0.35 & \textbf{29.14} / \textbf{0.90} / \textbf{0.24} & 28.33 / /0.88 / 0.27 \\
    Playroom & 30.04 / \textbf{0.91} / 0.32 & 30.17 / \textbf{0.91} / 0.33 & \textbf{30.37} / 0.90 / \textbf{0.31} & 22.98 / 0.80 / 0.50 & 21.67 / 0.78 / 0.43 & \textbf{29.66} / \textbf{0.90} / \textbf{0.25} & 29.56 / 0.89 / 0.26 \\
    \bottomrule
\end{tabular}
}
\caption{
   PSNR, SSIM, and LPIPS scores for Deep Blending~\cite{deepblending} scenes.
    }
\label{tab:db_all}
\end{table*}

\Cref{tab:mipnerf360_all,tab:db_all} summarize the error metrics collected for our evaluation of all considered techniques.
These include results for both Mip-NeRF360~\cite{mipnerf360} and Deep Blending~\cite{deepblending} scenes.
However, 3DGRT~\cite{3dgrt} is excluded from per-scene comparisons as these values are not reported in the original paper, and the code is not publicly available.

\vspace{33em}
\hfill

~
\vspace{42em}

\end{document}